\documentclass[10pt,a4paper,conference,composocconf]{IEEEtran}
\usepackage{cite}
\ifCLASSINFOpdf
   \usepackage[pdftex]{graphicx}
  \graphicspath{{img/}}
  \DeclareGraphicsExtensions{.pdf,.jpeg,.png}
\else
\fi

\usepackage{amsmath}
\usepackage{array}
\usepackage{url}

\usepackage[utf8]{inputenc}

\usepackage{amssymb} 
\usepackage{mathtools}
\usepackage{bm}
\usepackage{xspace}
\usepackage{microtype}
\usepackage{multicol}
\usepackage{booktabs}
\usepackage{nth}

\usepackage[table]{xcolor}

\usepackage{footnote}
\usepackage{afterpage}

\usepackage{multirow}
\usepackage{tikz}
\usepackage{pgfplots}
\usepgfplotslibrary{fillbetween}
\pgfplotsset{compat=newest}
\usetikzlibrary{patterns,
pgfplots.groupplots,fit,calc,positioning,shapes,spy}

\usepackage{tabularx}
\usepackage{booktabs}

\usepackage{varioref}
\usepackage[pagebackref=true,breaklinks=true,letterpaper=true,colorlinks,bookmarks=false,citecolor=green!80!black,linkcolor=red!80!black,urlcolor=blue]{hyperref}
\usepackage{cleveref}
\usepackage{subcaption}
\usepackage{csquotes}
\usepackage{siunitx}

\usepackage[inline]{enumitem}
\setlist*[enumerate]{label=(\arabic*)}

\newcommand{\onedot}{.\xspace}
\newcommand{\etal}[1]{#1~et~al\onedot}

\newcommand{\ie}{i.\,e.,\xspace}

\crefname{section}{Sec.}{Sections}
\crefname{figure}{Fig.}{Figure}
\crefname{table}{Tab.}{Table}
\crefname{equation}{Equ.}{Equation}

\definecolor{faublue}{RGB}{0,51,102}
\definecolor{darkgreen}{rgb}{0,0.6,0}
\definecolor{bblue}{HTML}{4F81BD}
\definecolor{rred}{HTML}{C0504D}
\definecolor{ggreen}{HTML}{9BBB59}
\definecolor{ppurple}{HTML}{9F4C7C}

\NewDocumentCommand{\rot}{O{60} O{1em} m}{\makebox[#2][l]{\rotatebox{#1}{#3}}}%
\NewDocumentCommand{\rotn}{O{90} O{1em} m}{\makebox[#2][l]{\rotatebox{#1}{#3}}}%
\NewDocumentCommand{\rotninety}{O{90} O{1em} m}{\makebox[#2][l]{\rotatebox{#1}{#3}}}%

\begin{document}
\title{ICDAR 2019 Competition on Image Retrieval for Historical Handwritten Documents}

\author{\IEEEauthorblockN{Vincent Christlein\IEEEauthorrefmark{1}, Anguelos Nicolaou\IEEEauthorrefmark{1},
Mathias Seuret\IEEEauthorrefmark{1}, Dominique Stutzmann\IEEEauthorrefmark{2}, Andreas Maier\IEEEauthorrefmark{1}}
\IEEEauthorblockA{
\IEEEauthorrefmark{1} Pattern Recognition Lab, Friedrich-Alexander-Universität Erlangen-Nürnberg,
91058 Erlangen, Germany\\
\{firstname.lastname\}@fau.de
}
\IEEEauthorblockA{
\IEEEauthorrefmark{2} Institut de Recherche et d'Histoire des Textes (Centre National de la Recherche Scientifique - UPR841), France\\
dominique.stutzmann@irht.cnrs.fr
}}

\maketitle

\begin{abstract}
This competition investigates the performance of large-scale retrieval of historical document images based on writing style. Based on large image data sets provided by cultural heritage institutions and digital libraries,
providing a total of 20\,000 document images representing about 10\,000 writers, divided in three types: writers of (i) manuscript books, (ii) letters, (iii) charters and legal documents. 
We focus on the task of automatic image retrieval to simulate common scenarios of humanities research, such as writer retrieval. 
The most teams submitted traditional methods not using deep learning techniques. 
The competition results show that a combination of methods is outperforming single methods. Furthermore, letters are much more difficult to retrieve than manuscripts.
\end{abstract}


\IEEEpeerreviewmaketitle

\section{Introduction} \label{sec:introduction}
Writer retrieval is an important task in the field of history, literary studies, and, particularly, paleography. Indeed, writer identification contributes to cultural studies in allowing to trace the intellectual activity of individuals and groups in past societies. Thanks to writer identification, one can ascribe some previously undated and anonymous writings to a particular person in connection with their known autographs. It helps us to understand their culture by identifying the texts they copied, either for themselves or in commission for patrons, and to trace the work methods of known authors (annotations, drafts, preliminary version, etc.) in order to have a better understanding of their philosophy and aims. By applying writer identification to the production of a group, such as a chancery, historians can also gain an idea of the inner organization and the relationships between individuals in this group. Therefore, the outcome may have direct impact on our knowledge of the past. Especially, in the age of mass digitization, a successful retrieval can assist humanists in their daily work. 

This competition is in line with previous ICDAR and ICFHR competitions on writer identification, but shares also some similarities to other retrieval tasks, such as word spotting. 
In word spotting, the challenge is to conduct an efficient image retrieval on small image patches.
While the amount of text is not an issue in this competition, the challenge is 
the extreme heterogeneity of the data material for finding relevant images. 

The last and so far only competition on historical writer identification~\cite{Fiel17ICDAR} consisted of 720 writers contributing five samples each, which resulted in 3\,600 samples.
In this competition, we increased the number of samples significantly. Therefore, we employed a semi-automatic procedure allowing us to gather 20\,000 document images from 10\,000 writers: 7\,500 pages from anonymous and isolated writers and 12\,500 pages for which a writer contributed two to five pages.
The task consists of finding all documents corresponding to a specific writer using a document from this writer as query. Those 7\,500 writers with a single page are used as distractor images. 

In this sense, this competition also shares similarities with the work of \etal{Wolf}~\cite{Wolf11}, 
who proposed a system to identify join candidates of the Cairo Genizah collection. 
Therefore, the authors used writer identification as a part of a larger framework to find these candidates in approximately 350\,000 fragments of mainly Jewish texts. 

The paper is organized as follows. In \cref{sec:data}, we give details about the data sources and the dataset splits. The submitted methods are explained in \cref{sec:methods}. \Cref{sec:evaluation} shows the results of the competition as well as additional experiments on data subsets. The paper is concluded in \cref{sec:conclusion}.

\section{Data}\label{sec:data}
The dataset is publicly available below: \url{https://doi.org/10.5281/zenodo.3262372}.

\subsection{Dataset sources}
\begin{figure*}[t]
    \centering
    \subcaptionbox{\label{fig:mss}}{\includegraphics[width=0.27\textwidth]{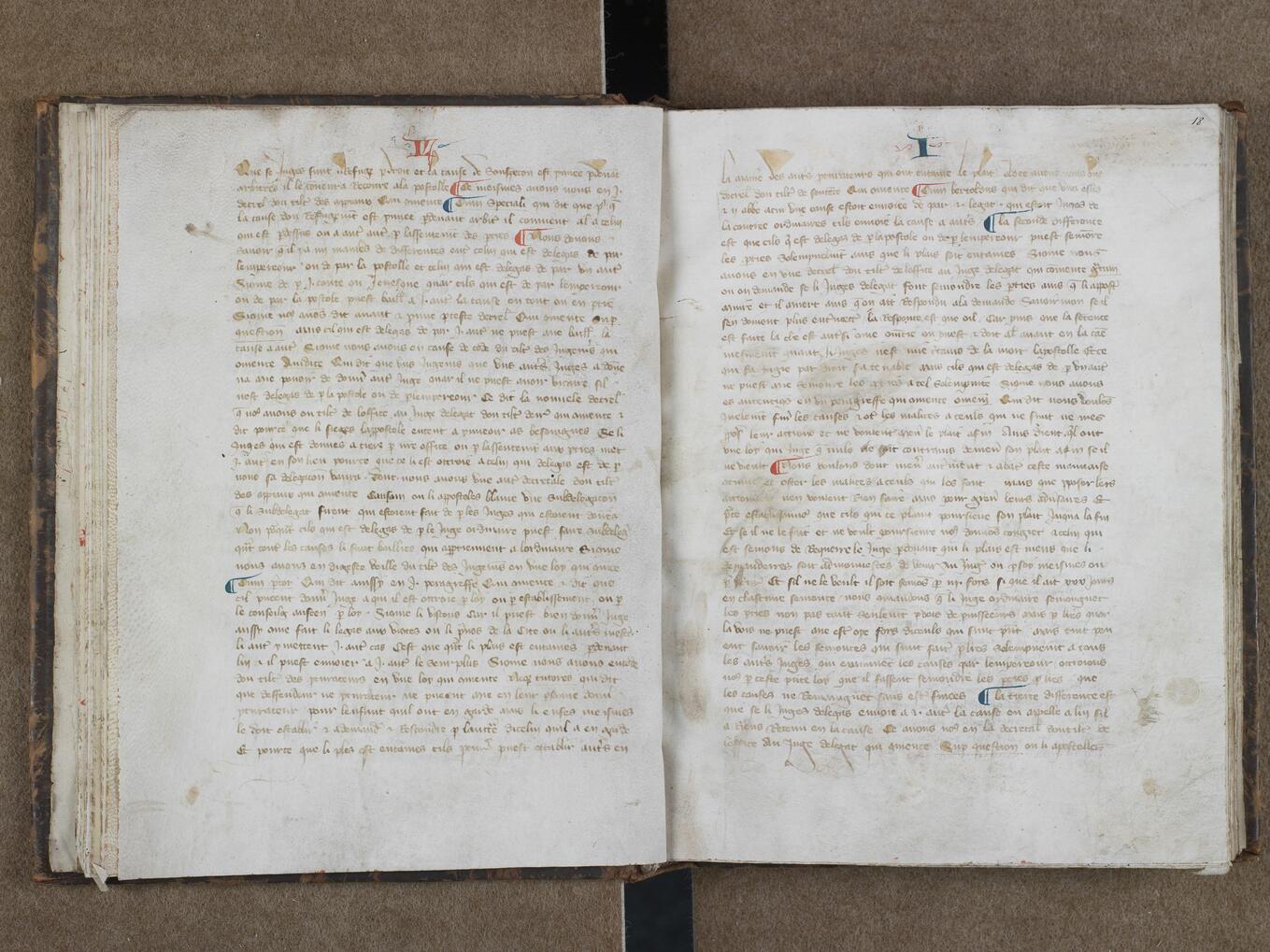}}
    \subcaptionbox{\label{fig:char}}{\includegraphics[width=0.28\textwidth]{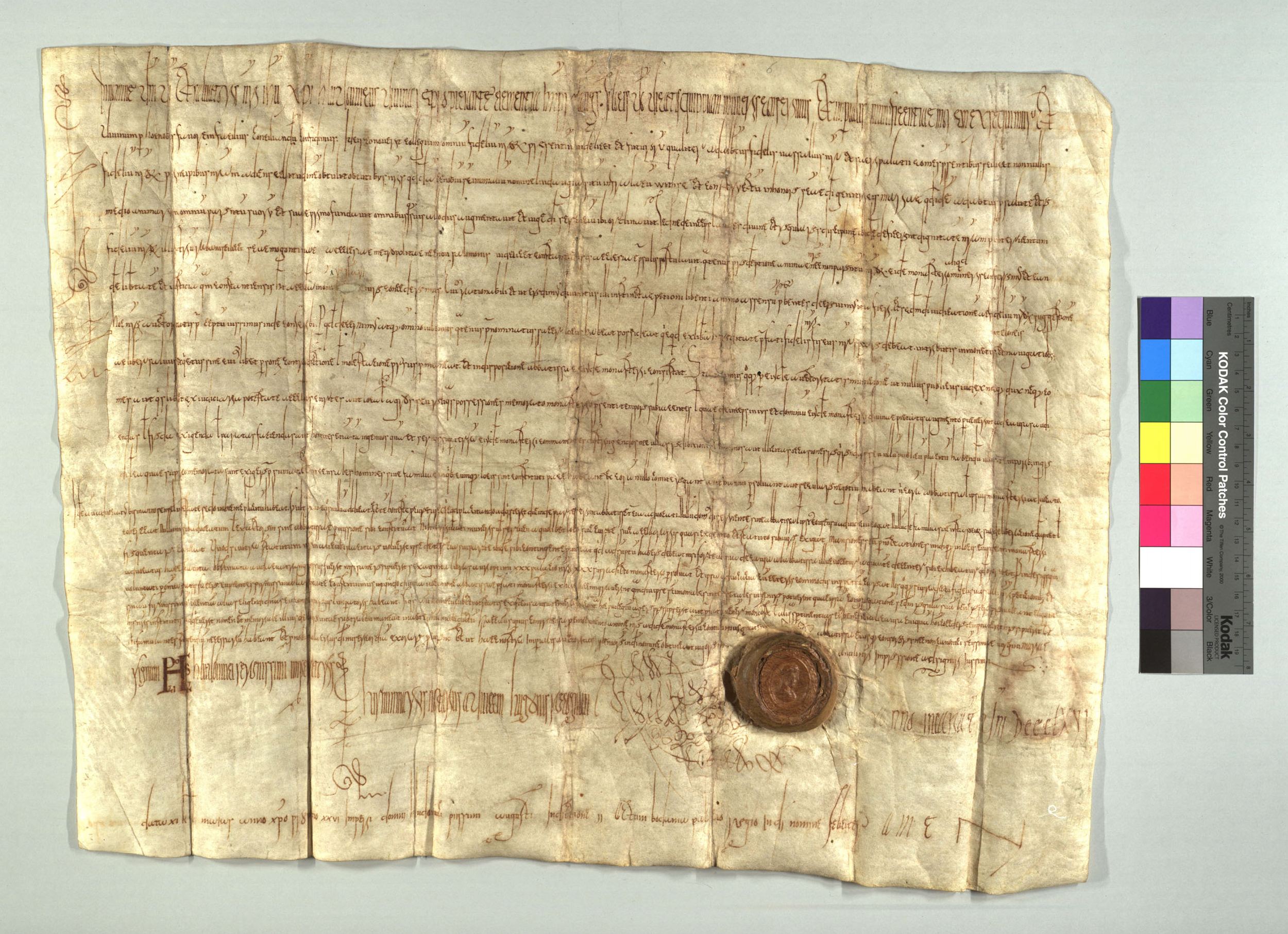}}
    \subcaptionbox{\label{fig:lA}}{\includegraphics[width=0.21\textwidth]{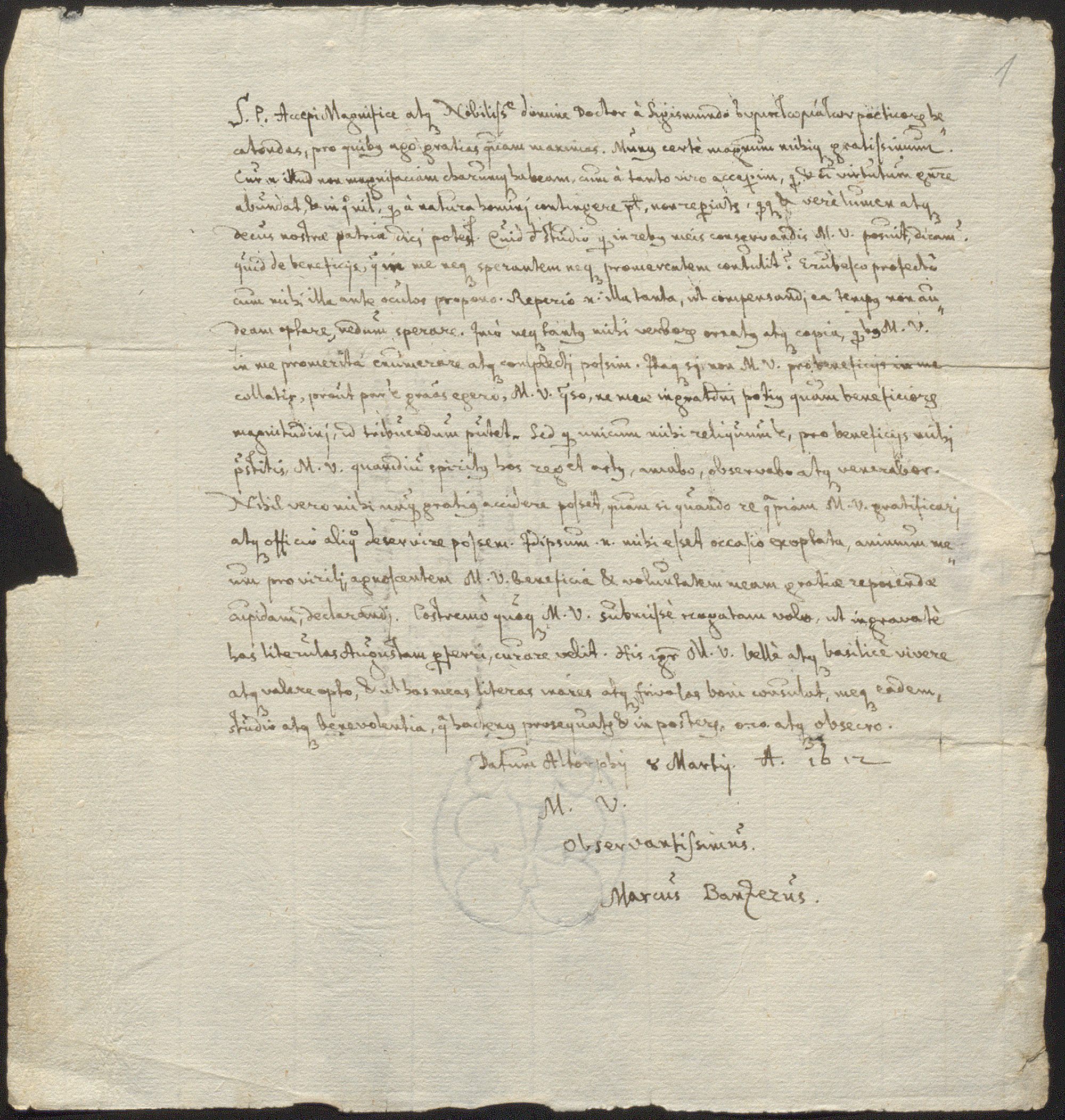}}
    \subcaptionbox{\label{fig:lB}}{\includegraphics[width=0.20\textwidth]{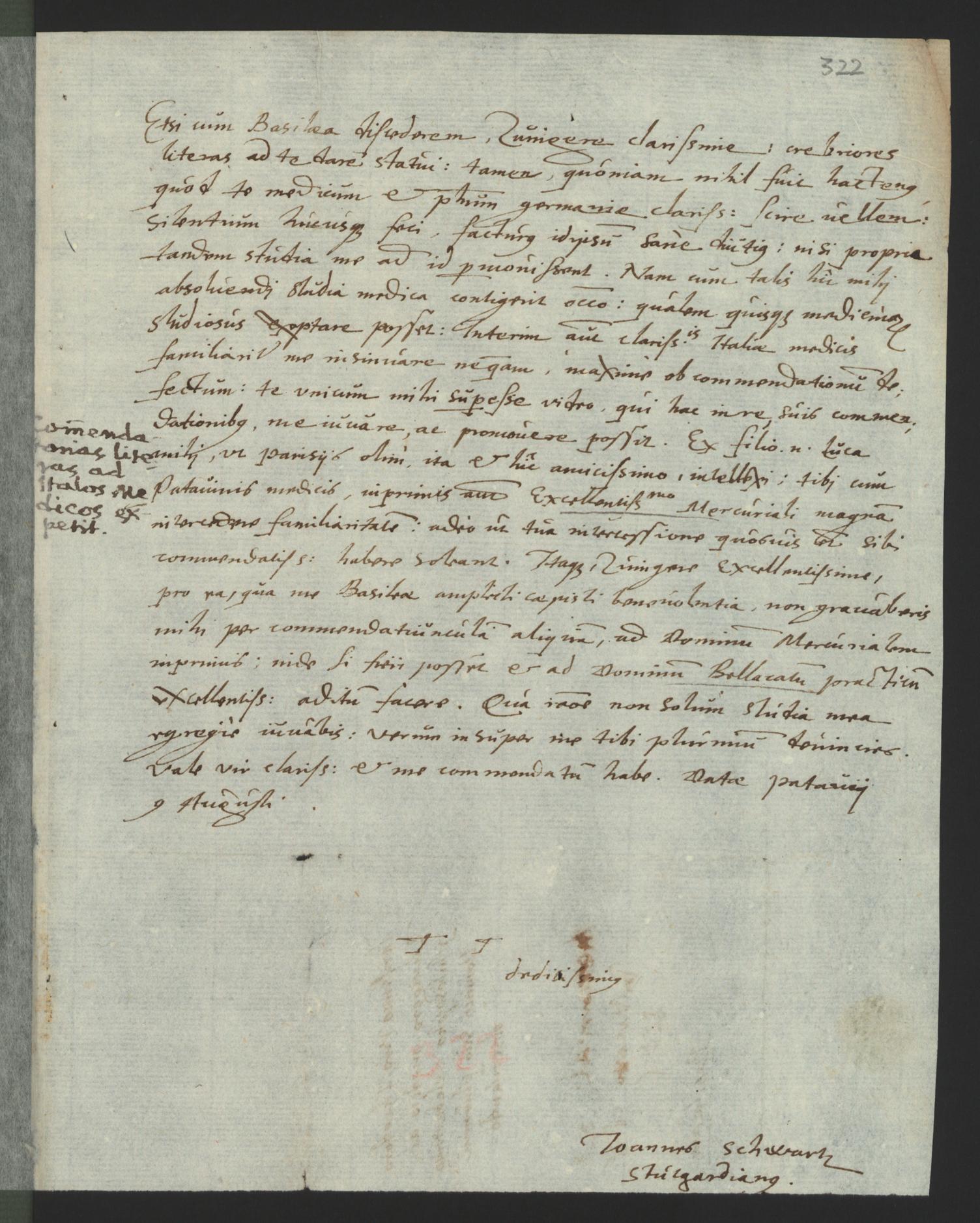}}
    \caption{Example images of the four subsets used in the dataset. 
    From left to right: 
    \subref{fig:mss} Saint-Omer, Bibliothèque d'agglomération, 545; BVMM, Saint Bertin;\protect\footnotemark
    \subref{fig:char} Haus-, Hof- und Staatsarchiv Lindau, Kanoniker AUR 839 IV 21; Monasterium.net;\protect\footnotemark
    \subref{fig:lA} Banzer, Marcus: Brief an Georg Remus [1612-03-08]; Die Briefsammlung des Nürnberger Arztes Christoph Jacob Trew (1695-1769); UB Erlangen-Nürnberg;\protect\footnotemark
    \subref{fig:lB} Schwartz, Johannes: Brief an Theodor Zwinger. Patavii, [16. Jh.]. Universitätsbibliothek Basel, Frey-Gryn Mscr II 26:Nr.322, \url{http://doi.org/10.7891/e-manuscripta-18277} / Public Domain Mark\protect\footnotemark
    }
    \label{fig:examples}
\end{figure*}

The competition dataset is composed of images from different sources and types. They can be grouped into: \begin{enumerate*}
    \item Manuscripts
    \item Letters
    \item Charters
\end{enumerate*}. Examples can be seen in \cref{fig:examples}.

\paragraph{Manuscripts}
The main focus of this corpus is the writers of books in the European Middle Ages, especially \nth{9} to \nth{15} century CE. The larger part of the corpus is anonymous. Indeed, few of the writer of this period signed their products and fewer are known by their names. In this part, given that paleographers’ attributions across books may be disputed, the organizers posit that consecutive pages in a homogeneous part of a book represent one particular writer.
\footnotetext[1]{\url{https://bvmm.irht.cnrs.fr/includes/php/rotation.php?vueId=1672647&niveauZoom=grand}}%
\footnotetext[2]{\url{http://monasterium.net/mom/AT-HHStA/LindauCan/AUR_839_IV_21/charter}}%
\footnotetext[3]{\url{https://nbn-resolving.org/urn:nbn:de:bvb:29-bv043513635-8}}%
\footnotetext[4]{\url{http://doi.org/10.7891/e-manuscripta-18277}}%
A smaller part of the corpus is composed by script samples from books that are believed or demonstrably known to have been written by the same individual, such as literary autographs. Concerning this subset and for the sake of homogeneity in the competition, the corpus comprises also five consecutive pages of each of the selected autograph books.

Most images are taken from IIIF compliant repositories allowing the use of images for scientific and teaching purposes. The main resources are Bibliothèque virtuelle des manuscrits médiévaux (BVMM)\footnote{\url{https://bvmm.irht.cnrs.fr/}} and Gallica.\footnote{\url{https://gallica.bnf.fr}} A full list of image providers can be found in \cref{tab:institutions}, individual city libraries, archives and shelfmarks are listed along the publicly available dataset.

\paragraph{Letters~A} From the University Library of the Friedrich-Alexander University Erlangen-Nürnberg, we thankfully obtained a subset of the digitized letter collection of the Nuremberg doctor Christoph Jacob Trew (1695--1769).\footnote{\url{http://digital.bib-bvb.de/R/5AL3NBRJYJV14LG6YC7RDNG4VHURY7SGHC4KASKKMDAH1LATRS-00090?func=collections-result&collection_id=2397}}

\paragraph{Letters B} Another set of mostly correspondences were used as single images of the test set. They come from the University Library Basel,\footnote{\url{https://www.unibas.ch}} licensed under public domain. The images are non-used (and non-published) ones of the ICDAR'17 writer identification competition~\cite{Fiel17ICDAR} but processed in a similar way.

\paragraph{Charters} For a large part of the single images of the test set, we used charters from Monasterium.net.\footnote{\url{https://www.monasterium.net/mom/home}}
To reduce the chance of obtaining multiple charters written by the same scribe, we used only charters where the date information was available. Then, we required that at least eight years lie in between any two images of one collection. 

\subsection{Dataset Splits}

\paragraph{Training and Validation}
For the training data, we suggested to use the ``ICDAR17 Historical-WI'' test dataset~\cite{Fiel17ICDAR}. We have not forbidden to use additional training data, such as the ICFHR2016 of ICDAR2017 ``Competition on the Classification of Medieval Handwriting in Latin Script''\cite{Cloppet17}.
This training data set was complemented by an additional corpus. It encompasses images from (i) \textit{Letters~A}, where each writer contributed one or three images; (ii) \textit{Manuscripts}, where each writer was represented by five consecutive images from a single book. 
In total, it contains 300 writers contributing one page, 100 writers contributing three pages, and 120 writers contributing five pages resulting in 1200 images of 520 writers.

\subsection{Testing}
The test data set contains 20\,000 images: 
About 7\,500 pages stem from isolated documents (partially anonymous writers, contributing one page each), and about 12\,500 pages are from writers that contributed three or five pages. Please see \cref{tab:institutions} for more details.
\begin{savenotes}
\begin{table*}[t]
    \centering
    \begin{tabular}{llrcr}
    \toprule
    Provider & City & nb writers & nb images/writer & images total\\
    \midrule
\textbf{Manuscripts} &	& \textbf{2027} & & \textbf{10\,135}\\\\
Bodleian Libr.\footnote{\url{https://www.bodleian.ox.ac.uk/}}	& Oxford & 9 & 5\\
BVMM\footnote{\url{https://bvmm.irht.cnrs.fr/}}	&  & 586 & 5\\
& Boulogne &	28\\
& Chantilly	& 30\\
& Nantes &	13\\
& Rennes & 16\\
& Saint-Omer & 363\\
& Toulouse	& 12\\
& $\leq10$ writers p.\ repository\footnote{Paris, Beaune, Angers, Metz, Auxerre, Versailles, Arras, Fécamp, Douai, etc.} &	124\\
Cambridge Dig. Libr.\footnote{\url{http://cudl.lib.cam.ac.uk/}}	& & 2 & 5\\
e-codices\footnote{\url{https://www.e-codices.unifr.ch}}	& Geneva & 2 & 5\\
Gallica\footnote{\url{https://gallica.bnf.fr}}	& & 1352 & 5\\
& Amiens & 14 & \\
& Paris &	1232 & \\
& Reims &	41 & \\
& Valenciennes	& 52 & \\
& $\leq8$ writers p. repository\footnote{Besançon, Bourges, Angers, Rouen, Louviers}	& 13 & \\
Harvard\footnote{\url{https://library.harvard.edu/}}	& & 19 & 5\\
Stanford\footnote{\url{http://library.stanford.edu/}}	& Baltimore (Walters) & 57 & 5\\
\midrule
\textbf{Letters~A} & & \textbf{831} & & \textbf{2655}\\ 
Univ.\ Library & Erlangen & \\
 -- Erlangen-Nürnberg & & 290 & 1\\
& & 170 & 3\\ 
& & 371 & 5\\ 
\midrule
\textbf{Letters B} & & \textbf{2052} & & \textbf{2052}\\
Univ.\ Library Basel & Basel & 2052 & 1\\
\midrule
\textbf{Charters} & & \textbf{5158} & & \textbf{5158}\\
Monasterium\footnote{\url{https://www.monasterium.net/mom/home}} & & 5158 & 1\\
\midrule
Total & & 10\,068 & & 20\,000\\
\bottomrule
    \end{tabular}
    \caption{Image providers and number of writers used in the test dataset.}
    \label{tab:institutions}
\end{table*}
\end{savenotes}

\subsection{Preprocessing}
We cropped all selected images slightly at the borders (42 pixels each) of the image in order to avoid that the participants can directly search for the image by using byte hashing techniques. 
For the training data, the Manuscript images were not resized while the ones of the Letters~A dataset were resized to a fixed width of 2000 pixels and all images (Letters~A + Manuscripts) were JPEG compressed with a quality factor of 81, which is fine for text documents. 

For the test set, we had to downscale all images to reduce the download size of the full dataset. Thus, we resized every image independent of their source such that the larger dimension (height or width) becomes 2000 pixel.
Original JPEG images (Manuscripts, Letters B, Charters) were saved again in high quality (JPEG quality 96), while images of Letters~A (where we obtained the original tif sources) were saved with a low quality factor (81), in accordance to the terms of use of the library (same image quality as the online resources). 

\section{Methods}\label{sec:methods}

\subsection{Baseline}
In total, we obtained nine submissions from three participants. Additionally, we evaluated the method of \etal{Nicolaou}~\cite{Nicolaou15ICDAR} in two variants as our baselines. One was designed with retrieval in mind and one with classification. In the following sections, the submissions are explained briefly.
In both variants, a special variation of Local Binary Patterns (LBP), designed for document image analysis, are computed for radii 1 to 12.
The patterns of each radius are globally pooled into a histogram of 256 bins and normalized.
The histograms are then concatenated, they embed the texture of each text image into a 3072-dimensional vector.
The embedded samples are then mapped to $\mathbb{R}^{200}$ along the principal components of either (a) an external data-set (here: the ICDAR19 training dataset) representing a classification scenario or (b) the test-set (here: the ICDAR19 test set) representing a retrieval scenario where we have access to all embeddings of the test set.
The origin of the principal components, is the only difference between the classification and the retrieval variants of the baseline.
As a final step the Hellinger kernel is applied on the vectors and afterwards normalized by their $\ell^2$ norm. 
The distances are computed by the Manhatten distance. 
For more details refer to~\cite{Nicolaou15ICDAR}.
The exact implementation employed is available online.\footnote{\url{https://github.com/anguelos/wi19_evaluate/tree/master/srslbp}}

\subsection{South China University of Technology (SCUT)}
Songxuan Lai and Lianwen Jin from the South China University of Technology (SCUT) submitted in total three methods. 
Because of the complex layout and background of the dataset, the participants first used a deep Unet for text binarization. They used the implementation of Mikhail Masyagin,\footnote{\url{https://github.com/masyagin1998/robin}} but trained their own model using a larger dataset. After binarization, they performed a page-level rotation correction step based on the line projection method to make the text more horizontal. Then, based on the corrected gray images and binary images, they extracted two fundamental features.

\paragraph{SIFT}
The first one is SIFT~\cite{Lowe04} with a key modification: we always set the angle to zero, similar to the work of \etal{Christlein}~\cite{Christlein17PR}. This slight modification is aimed to capture local structures with different orientations. Besides, they only extracted SIFT features from the foreground, \ie text, by using the binary images as masks. Thereafter, they used the Fisher Vector~\cite{Perronnin07} to encode the extracted SIFT features. The codebook was trained using the ICDAR17 historical writer identification dataset~\cite{Fiel17ICDAR}.

\paragraph{Pathlet}
The second submitted method uses their newly proposed pathlet feature~\cite{Lai19}, which is aimed to capture useful information such as curvature and slant from the contours. Pathlets were extracted from the binary images and encoded using VLAD encoding~\cite{Jegou12ALI}. The codebook was also trained using the ICDAR17 dataset~\cite{Fiel17ICDAR}.

\paragraph{SIFT + Pathlet}
The above two feature vectors were projected onto a subspace of lower dimensionality and then concatenated. The projection matrices were obtained by performing SVD on the feature matrices of the ICDAR17 dataset~\cite{Fiel17ICDAR}. The concatenated vectors were further power-normalized and l2-normalized and used as the final feature vectors.

Finally, Euclidean distances of the global descriptors were computed.

\subsection{University of Groningen}
Sheng He from the University of Groningen submitted the outcomes of the following five methods:
\begin{enumerate*}
    \item \emph{Hinge},
    \item \emph{Co-Hinge},
    \item \emph{QuadHinge},
    \item \emph{Quill},
    \item \emph{TCC}.
\end{enumerate*}
The methods do not have a training stage involved and work purely on the basis of  binary images. Detailed descriptions of the single methods can be found in~\cite{He17BOM}. All methods were applied on binarized images using Otsu's binarization method and compared using $\chi^2$ distance.

\subsection{University of Tébessa}
Abdeljalil Gattal, Chawki Djeddi and Faycel Abbas from the University of Tébessa submitted the following method.
The different configurations of oriented Basic Image Features (oBIFs) columns histograms~\cite{Abdeljalil16,Abdeljalil18} are extracted from historical document samples and concatenated for generating a feature vector. Classification is carried out using Euclidean distance in these experiments while the oBIF parameter $\epsilon$ is fixed to $0.1$. The technique also does not require any preprocessing and the features are directly extracted from the complete images of handwriting, \ie no binarization is needed.

\section{Evaluation}
\label{sec:evaluation}

\subsection{Error Metrics}
\label{sct:error-metrics}
Each participant was required to hand in a $20000\times20000$ distance matrix. We evaluated the results of the participants in a leave-one-image-out cross-validation manner. This means that every image of the test set will be used as query for which the other test images are ranked. The metrics are then averaged over all queries.
Our test-set is unbalanced class-wise. Queries can have from four samples down-to no samples of the same class to be retrieved. 
Several different metrics will be evaluated and reported. 
The metric that decides upon winning is mean Average Precision (mAP) which is estimated on the distance matrix, which is provided by the participants. 
Mean Average Precision is computed as follows. For each query, the average precision (AP) is computed. Therefore, the precision over all ranks $r$ is computed. Given the retrieved list of size $S$ in which $R$ documents are relevant for the query $q$, then the AP is computed as:
\begin{equation}
    \mathrm{AP}_q = \frac{1}{R} \sum_{r=1}^S \mathrm{Pr}_q (r) \cdot \text{rel}_q(r) \;,
\end{equation}
where $\mathrm{Pr}_q(r)$ is the precision at rank $r$ and $\text{rel}_q(r)$ is an indication function returning $1$ if the document at rank $r$ is relevant and $0$ otherwise. 
The mAP also represents the area under the curve of the precision recall curve, \ie the curve that is created when plotting the precision as a function of recall. An evaluation system, which computed the mAP was provided to the particpants below \url{https://github.com/anguelos/wi19_evaluate}. 
Furthermore, we report the Top-1 accuracy, \ie the average precision at rank $1$.

\subsection{Results}
We first evaluate the full test dataset. Afterwards, we evaluate on different subsets to have a better picture on their individual influence. 

\subsubsection{Full Test Dataset}

\begin{table}[t]
    \centering
        \caption{Competition results for the full test set.}
    \label{tab:results}
    \begin{tabular}{llcc}
    \toprule
    & Method &  Accuracy [\%] & mAP [\%]\\
    \midrule
    \multirow{2}{*}{Baseline}
    & SRS-LBP (a) Classification         & 92.2    & 84.0  \\
    & SRS-LBP (b) Retrieval         &  93.1     & 86.8 \\
    \midrule
    \multirow{3}{*}{SCUT}
    & SIFT          &    96.6    & 90.6 \\ 
    & Pathlet       &  96.0   & 89.8  \\
    & SIFT + Pathlet &  97.4   
    & \textbf{92.5} \\  
    \midrule
    \multirow{5}{*}{Groningen}
    & Hinge          & 88.4 & 75.6\\
    & Co-Hinge       & 92.9 & 84.5\\
    & QuadHinge      & 91.3 & 80.2\\
    & Quill          & 88.3 & 76.0\\
    & TCC            & 89.7 & 79.0\\
    \midrule
    Tébessa & oBIFs & 92.7  &  84.6\\
    \bottomrule     
    \end{tabular}
\end{table}
\Cref{tab:results} shows that the team with the highest mAP is SCUT. The combination of Fisher Vector-encoded SIFT descriptors and VLAD-encoded Pathlet features surpassed all other methods. Interestingly, the traditional bag-of-words method using SIFT and Fisher Vectors achieves the second highest results. 

The Co-Hinge method of Groniningen and the oBIFs method of Tébessa achieve quite similar results. This is in accordance to the results of the ICDAR'17 Historical-WI competition~\cite{Fiel17ICDAR}, where these two teams made the first two places. Noteworthy, our baseline method is superior to both methods by a small margin. 

\subsubsection{Subset Results}

\begin{table*}[t]
    \centering
    \caption{Competition results in mAP for different subsets.}
    \label{tab:results_subsets}
    \begin{tabular}{llccccc}
    \toprule
    & Method &  MSS & MSS + Chars & Letters~A & Letters~A + B & Full \\
    \midrule
    \multirow{2}{*}{Baseline}
    & SRS-LBP Classification    & 93.0 & 93.0 & 45.8 & 45.4 & 84.0\\
    & SRS-LBP Retrieval         & 96.1 & 96.1 & 47.4 & 47.1 & 86.8\\
    \midrule
    \multirow{3}{*}{SCUT}
    & SIFT      & 95.4 & 95.4 & 70.2 & 70.1 & 90.6\\
    & Pathlet   & 94.7 & 94.7 & 70.0 & 69.2 & 89.8\\
    & SIFT + Pathlet  & \textbf{97.0} & \textbf{97.0} & \textbf{74.0} & \textbf{73.8} & \textbf{92.5}\\  
    \midrule
    \multirow{5}{*}{Groningen} 
    & Hinge          &  84.2 & 84.0 & 40.5 & 40.0 & 75.6\\
    & Co-Hinge       &  92.7 & 92.7 & 50.5 & 50.0 & 84.5\\
    & QuadHinge      &  87.9 & 87.8 & 48.0 & 47.7 & 80.2\\
    & Quill          &  85.6 & 85.4 & 36.6 & 36.3 & 76.0\\
    & TCC            &  88.0 & 87.9 & 41.6 & 41.2 & 79.0\\
    \midrule
    Tébessa & oBIFs &  93.2 & 93.2 & 48.4 & 47.8 & 84.6\\
    \bottomrule     
    \end{tabular}
\end{table*}
We evaluated four different dataset combinations to obtain a better picture on their individual influences (table abreviations in brackets):
\begin{enumerate*}
\item Manuscripts (MSS)
\item Manuscripts + Charters (MSS + Chars)
\item Letters~A 
\item Letters~A + Letters B (Letters~A + B)
\end{enumerate*}

The results are shown in \cref{tab:results_subsets}. 
Comparing the results for the Manuscripts subset and Manuscripts + Charters subset, we see that there is no difference. 
That means that using images of charters as distractor images for Manuscript images is not very useful. Probably, the two subsets were too different from each other resulting in no confusion. For future competitions, these images could be omitted. 

Conversely, the Letters B subset worked as a distractor subset for Letters~A. However, the recognition results were only slightly reduced. Overall, the letters seem to be much more challenging than the manuscript images. Interestingly, our baseline method ranks second on the Manuscripts subsets. 

\section{Conclusion}\label{sec:conclusion}
This competition focused on large-scale historical document image retrieval. In particular, we were interested in finding the same scribe among a large image dataset. 
Therefore, we created a test set containing 20\,000 images provided by different cultural heritage institutions and digital libraries. The dataset consists of more than 10\,000 writers of different document types (manuscripts, letters, charters), which show a wide variation in writing style and document appearance. 

The winner of this competition is the team of the South China University of Technology, whose submission of the combination of SIFT and their newly developed Pathlet features obtained the overall best results.  
Our evaluation shows that historical data is still quite challenging, obtaining an mAP of about \SI{93}{\percent} in comparison to contemporary data which achieves recognition rates beyond \SI{99}{\percent}~\cite{Christlein19}. 
Interestingly, none of the methods rely on neural networks. Only one method uses a convolutional neural network for image binarization. 
We believe that in the future, more neural network-based approaches will dominate also retrieval-based tasks, such as writer identification/retrieval.
A perspective to make the task even closer to Humanities research is to use the full resolution images but selecting text regions of different (random) sizes. 
This would be parallel to identifying the writer of an annotation in a manuscript with their known autographs.

\bibliographystyle{IEEEtran}
{\small
\bibliography{references}
}
\end{document}